\pdfoutput=1
\RequirePackage{rotating}
\documentclass[11pt]{article}

\usepackage{acl}

\usepackage{times}
\usepackage{latexsym}
\usepackage{tabularx}
\usepackage{multicol}
\usepackage{graphicx}
\usepackage{arydshln}
\usepackage{multirow}
\usepackage{array}
\usepackage{colortbl}
\usepackage[inline]{enumitem}
\usepackage{multirow}
\usepackage{graphicx}
\usepackage{longtable}
\usepackage{tikz}
\usepackage{cleveref}
\usepackage{rotating}
\usepackage[T1]{fontenc}
\definecolor{Gray}{gray}{0.96}
\newcolumntype{G}{>{\columncolor{Gray}}c}

\usepackage[utf8]{inputenc}

\usepackage{microtype}
\usepackage{mathrsfs}
\title{WLASL-LEX: a Dataset for Recognising Phonological Properties in American Sign Language}

\author{Federico Tavella \and Viktor Schlegel \and Marta Romeo \\ {\bf Aphrodite Galata} \and {\bf Angelo Cangelosi} \\
\texttt{\{name.surname\}@manchester.ac.uk} \\
         Department of Computer Science, The University of Manchester}

\begin{document}
\maketitle
\begin{abstract}
Signed Language Processing (SLP) concerns the automated processing of signed languages, the main means of communication of Deaf and hearing impaired individuals. SLP features many different tasks, ranging from sign recognition to translation and production of signed speech, but it has been overlooked by the NLP community thus far.
In this paper, we bring attention to the task of modelling the phonology of sign languages. We leverage existing resources to construct a large-scale dataset of American Sign Language signs annotated with six different phonological properties. We then conduct an extensive empirical study to investigate whether data-driven end-to-end and feature-based approaches can be optimised to automatically recognise these properties. We find that, despite the inherent challenges of the task, graph-based neural networks that operate over skeleton features extracted from raw videos are able to succeed at the task to a varying degree. Most importantly, we show that this performance pertains even on signs unobserved during training. 


\end{abstract}

\section{Introduction}
Around 200 languages in the world are signed rather than spoken, featuring their own vocabulary and grammatical structures. For example the American Sign Language (ASL) is not a mere translation of English into signs and is unrelated to the British Sign Language (BSL). Their non-textual nature introduces many challenges to their automated processing, compared with purely textual NLP. Research on Sign Language Processing (SLP) encompasses tasks such as sign language detection, i.e. recognising if and which signed language is performed \cite{Moryossef2020Real-TimeEstimation} and sign language recognition (SLR) \cite{Koller2020QuantitativeRecognition}, i.e. the identification of signs either in isolation or in continuous speech. Other tasks concern the translation from signed to spoken (or written) \cite{Camgoz2018NeuralTranslation} language or the production of signs from text \cite{Rastgoo2021SignSurvey}. With the recent success of deep learning-based approaches in computer vision (CV), as well as advancements in ---from the CV perspective---related tasks of action and gesture recognition \cite{Asadi-Aghbolaghi2017ASequences}, SLP is gaining more attention in the CV community \cite{Zheng2017RecentRecognition}.

\begin{figure}[t]
    \resizebox{.95\columnwidth}{!}{\input{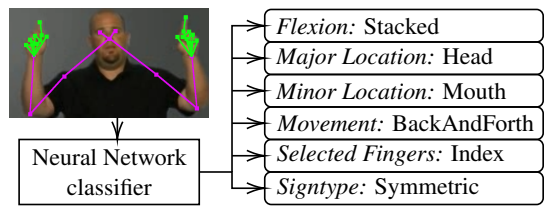}}
    \caption{We annotate ASL sign videos with their corresponding phonological information and skeleton features of the speakers, and train neural networks to recognise the former from the latter.}
    \label{fig:my_label}
\end{figure}

Due to the complexity of the tasks, some recent approaches to various SLP tasks implicitly rely on \emph{phonological} features \cite{Tornay2021ExplainableAssessment,Metaxas2018ScalableCorpora,Gebre2013AutomaticIdentification, tavella2021phonology}. Surprisingly, however, little work has been carried out on explicitly modelling the phonology of signed languages. This presents a timely opportunity to investigate signed languages from the perspective of computational linguistics \cite{Yin2021IncludingProcessing}. In the context of signed languages, phonology typically distinguishes between manual features, such as usage, position and movement of hands and fingers, and non-manual features, such as facial expressions. Sign language phonology is a matured field with well-developed theoretical frameworks \cite{Liddell1989AmericanBase,Fenlon2017SignPhonology,Sandler2012TheLanguages}.
These phonological features, or \emph{phonemes}, are drawn from a fixed inventory of possible configurations which is typically much smaller than the vocabulary of signed languages  \cite{Borg2020Phonologically-MeaningfulRecognition}. For example, there is only a limited number of fingers that can be used to perform a sign due to anatomical constraints. Hence, different signs share phonological properties and well performing classifiers can be used to predict those properties for signs unseen during training. This potentially holds even across different languages, because, while different languages may dictate different combinations of phonemes, there are also significant overlaps \cite{Tornay2020TowardsRecognition}.

Finally, these phonological properties have a strong discriminatory power when determining signs. For example, in ASL-Lex \cite{Caselli2017}, a lexicon which also captures phonology information, the authors report that more than 50\% of its 994 described signs have a unique combination of only six phonological properties and more than 80\% of the signs share their combination with at most two other signs. By relying on this phonological information from resources such as ASL-Lex, many signs can be uniquely determined. This means that well performing classifiers can leverage this information to predict signs without having encountered them during training. This is a capability that current data-driven approaches to SLR lack by design \cite{Koller2020QuantitativeRecognition}.
Thus, in combination, mature approaches to phonology recognition can facilitate the development of sign language resources, for example by providing first-pass silver annotations for new sign languages based on their phonological properties. This is an important task for both documenting low-resource sign languages as well as rapid developing of large-scale datasets, and for fully harnessing data-driven CV approaches.

To spur research in this direction, we extend the preliminary work by \newcite{tavella2021phonology} and introduce the task of Phonological Property Recognition (PPR). More specifically, with this paper, we contribute 
\begin{enumerate*}[label=\emph{(\roman*)}]
    \item WLASLLex2001, a large-scale, automatically constructed PPR dataset,
    \item an analysis of the dataset quality, and
    \item an empirical study of the performance of different deep-learning based baselines thereon.
\end{enumerate*}

\section{Methodology}
We address PPR as a classification problem based on features extracted from videos of people speaking SL. Although manual annotation approaches are widely adopted, these are time consuming and require expert knowledge. Instead, we rely on automated dataset construction. On a high level, we cross-reference a large-scale ASL SLR dataset with an ASL Lexicon and annotate videos of signs with their corresponding phonological properties. We then extract skeletal features, by taking advantage of pre-trained deep models from the computer vision community~\cite{rong2021frankmocap, WangSCJDZLMTWLX19}. 
Finally, we train several deep models to classify them as phonological classes.

\subsection{Dataset construction}
As previously mentioned, ASL-Lex~\cite{Caselli2017} contains phonological features of American Sign Language, such as where the sign is executed, the movement performed by the hand and the number of hands and fingers involved.
The latter properties were coded by 3 ASL-versed people.
In our work, we are interested in recognising phonological properties from videos of people speaking ASL. Consequently, we aim to construct a dataset, suitable for supervised learning, containing videos labelled with six phonological properties. Specifically, we choose the manual properties with the strongest discriminatory power to determine signs based on their configuration~\cite{Caselli2017}:
\begin{enumerate}[label=(\roman*)]
    \item \emph{flexion:} aperture of the selected fingers of the dominant hand at sign onset,
    \item \emph{major location:} general location of the dominant hand at sign onset,
    \item \emph{minor location:} specific location of the dominant hand at sign onset,
    \item \emph{movement:} the first movement path of the sign, 
    \item \emph{selected fingers:} fingers that are moving or are foregrounded during that movement, and 
    \item \emph{sign type:} symmetry of the hands according to \newcite{battison1978lexical}.
\end{enumerate}     
A detailed description of all the properties is provided in the appendix. 

One of the limitations of ASL-Lex is the small number of examples and lack of variety: its first iteration (ASL-Lex 1.0) contains less than 1000 videos, all signed by the same person. While sufficient for educational purposes, these videos are of limited suitability for developing robust classifiers that can capture the diversity of ASL speakers~\cite{Yin2021IncludingProcessing}. To this end, we source videos from WLASL~\cite{li2020word} (Word Level-ASL), one of the largest available SL datasets, featuring more than 2000 glosses demonstrated by over 100 people, for a total of more than 20000 videos. 
Each sign is performed by at least 3 different signers, which implies greater variability compared to having one gloss performed by only one user. 
By cross referencing ASL-Lex and WLASL2000 based on corresponding glosses, we can increase the number of samples available to train our models.

Finally, to leverage state of the art SLR architectures that operate over structured input, we enrich each raw video with its extracted keypoints that represent the joints of the speaker. To do so, we use two pretrained models, FrankMocap~\cite{rong2021frankmocap} and HRNet~\cite{WangSCJDZLMTWLX19}. 
While these tracking algorithms follow different paradigms, the former extracting 3D coordinates based on a predicted human model and the latter predicting keypoints as coordinates from videos directly, they produce similar outputs. An important distinction is that while FrankMocap estimates the 3D keypoints, HRNet outputs 2D keypoints with associated prediction confidence scores. We use these different models to explore whether different tracking algorithms affect the recognition of phonological classes. 
We select a subset of features of the upper body, namely: nose, eyes, shoulders, elbows, wrists, thumbs and first/last knuckles of the fingers. These manual features were determined to be the most informative while performing sign language recognition~\cite{Jiang2021SkeletonRecognition}.

Our final dataset, WLASL-Lex2001 (WLASL2000 + ASL-Lex 1.0), is composed of 10017 videos corresponding to 800 glosses, 3D skeletons ($x$, $y$, $z$ from FrankMocap and $x$, $y$ and $score$ from HRNet) labelled with their phonological properties.
A characteristic of this dataset is that it follows a long tailed distribution. Due to the nature of language, some phonological properties are more common than others, which means that some classes are more represented than others. On the one hand, the training setup for our models should take this factor into account, but on the other hand, the advantage of training over phonological classes instead of glosses is that different glosses can share phonological classes.

\subsection{Models} 
To estimate the complexity of the dataset, we use the majority-class baseline and the Multi-Layer Perceptron (MLP) as basic deep models. We further use Long Short-Term Memory (LSTM) and Gated Recurrent Units (GRU) as models capable of capturing the temporal component of videos. As state-of-the-art SLP architectures that have been used to perform SLR, we use the I3D 3D Convolutional Neural Network~\cite{i3d,li2020word} able to learn from raw videos, and the Spatio-Temporal Graph Convolutional Network (STGCN)~\cite{Jiang2021SkeletonRecognition} that captures both spatial and temporal components from the extracted keypoints.

\begin{table*}[!t]
    \centering
    
    \resizebox{.99\textwidth}{!}{%
    \begin{tabular}{p{0.05cm} l | c G c G c G c G c G c G}
    &  & \multicolumn{2}{c}{\textbf{\textsc{Flexion}}} &  \multicolumn{2}{c}{\textbf{\textsc{MajLocation}}} &  \multicolumn{2}{c}{\textbf{\textsc{MinLocation}}}  &  \multicolumn{2}{c}{\textbf{\textsc{Movement}}}  &  \multicolumn{2}{c}{\textbf{\textsc{Fingers}}}  &  \multicolumn{2}{c}{\textbf{\textsc{Signtype}}}  \\
    &  & $A$ & $\overline{A}$ & $A$ & $\overline{A}$ & $A$ & $\overline{A}$ & $A$ & $\overline{A}$ & $A$ & $\overline{A}$ & $A$ & $\overline{A}$  \\
        \hline
        \hline
        \multirow{7}{*}{\rotatebox[origin=c]{90}{\emph{Phoneme}}} &
Baseline & $50.3$ & $11.1$ & $34.4$ & $20.0$ & $33.9$ & $3.1$ & $35.5$ & $16.7$ & $48.2$ & $11.1$ & $39.3$ & $20$ \\
& MLP$_{H}$          & $44.1 \pm 2.5$ &  $11.1$ &       $70.3 \pm 2.3$ &     $64.0$ &              $51.6 \pm 2.5$ &     $28.2$ &              $34.5 \pm 2.4$ &  $18.7$ &         $59.4 \pm 2.5$ &       $25.0$ &                $73.9 \pm 2.2$ &  $52.6$ \\
& MLP$_{F}$          & $50.3 \pm 2.5$ &  $11.1$ &        $57.8 \pm 2.5$ &     $46.8$ &              $34.3 \pm 2.4$ &     $9.1$ &               $34.3 \pm 2.4$ &  $18.7$ &         $43.4 \pm 2.5$ &       $12.9$ &                $67.0 \pm 2.4$ &  $42.8$  \\
& RNN$_{H}$           & $49.0 \pm 2.5$ &  $30.0$ &        $75.8 \pm 2.2$ &     $72.4$ &              $64.3 \pm 2.4$ &     $46.0$ &              $35.1 \pm 2.4$ &  $29.5$ &         $71.0 \pm 2.3$ &       $46.5$ &                $78.7 \pm 2.1$ &  $58.8$ \\
& RNN$_{F}$         & $50.3 \pm 2.5$ &  $11.1$ &        $64.6 \pm 2.4$ &     $54.2$ &              $30.3 \pm 2.3$ &     $4.0$ &               $35.4 \pm 2.4$ &  $18.1$ &         $46.5 \pm 2.5$ &       $12.4$ &                $70.9 \pm 2.3$ &  $46.8$   \\
& STGCN$_{H}$           & $\mathbf{62.3 \pm 2.4}$ &  $\mathbf{45.0}$ &        $\mathbf{83.2 \pm 1.9}$ &     $\mathbf{78.6}$ &              $\mathbf{74.5 \pm 2.2}$ &     $\mathbf{63.5}$ &              $\mathbf{63.6 \pm 2.4}$ &  $\mathbf{58.2}$ &         $\mathbf{73.8 \pm 2.2}$ &       $\mathbf{56.0}$ &                $\mathbf{84.5 \pm 1.8}$ &  $\mathbf{69.6}$  \\
& STGCN$_{F}$           & $43.4 \pm 2.5$ &  $20.8$ &        $70.5 \pm 2.3$ &     $62.1$ &              $53.0 \pm 2.5$ &     $40.0$ &              $45.7 \pm 2.5$ &  $37.8$ &         $63.1 \pm 2.4$ &       $32.8$ &                $73.0 \pm 2.2$ &  $53.1$  \\
\hdashline
& 3DCNN & $46.5 \pm 2.5$ &  $13.2$ &        $64.3 \pm 2.4$ &     $55.2$ &              $42.3 \pm 2.5$ &     $18.6$ &              $32.9 \pm 2.4$ &  $20.8$ &         $47.5 \pm 2.5$ &       $14.5$ &                $69.5 \pm 2.3$ &  $44.8$ \\
        \hline
        \hline
        \multirow{6}{*}{\rotatebox[origin=c]{90}{\emph{Gloss}}} &
Baseline & $\mathbf{53.1}$ & $11.1$ & $35.7$ & $20.0$ & $42.0$ & $5.0$ & $35.2$ & $16.7$ & $47.4$ & $12.5$ & $38.3$ & $20.0$ \\
& MLP$_{H}$          & $44.6 \pm 2.5$ &  $15.5$ &        $68.1 \pm 2.3$ &     $56.6$ &              $47.3 \pm 2.5$ &     $19.7$ &              $28.4 \pm 2.2$ &  $19.8$ &         $56.2 \pm 2.5$ &       $22.9$ &                $75.3 \pm 2.2$ &  $50.7$  \\
& MLP$_{F}$          & $52.8 \pm 2.5$ &  $11.1$ &        $56.6 \pm 2.5$ &     $42.9$ &              $38.3 \pm 2.4$ &     $10.7$ &              $37.1 \pm 2.4$ &  $21.7$ &         $39.3 \pm 2.5$ &       $12.5$ &                $68.4 \pm 2.4$ &  $41.2$  \\
& RNN$_{H}$           & $39.6 \pm 2.5$ &  $18.0$ &        $72.8 \pm 2.2$ &     $67.3$ &              $49.3 \pm 2.5$ &     $26.3$ &              $32.2 \pm 2.3$ &  $24.9$ &         $60.7 \pm 2.5$ &       $32.5$ &                $75.4 \pm 2.2$ &  $53.5$ \\
& RNN$_{F}$          & $53.0 \pm 2.5$ &  $11.1$ &        $64.1 \pm 2.4$ &     $52.6$ &              $44.4 \pm 2.4$ &     $17.8$ &              $36.7 \pm 2.4$ &  $20.1$ &         $27.3 \pm 2.3$ &       $12.7$ &                $72.0 \pm 2.3$ &  $46.9$ \\
& STGCN$_{H}$      &    $49.1 \pm 2.5$ &  $\mathbf{21.6}$ &        $\mathbf{77.3 \pm 2.1}$ &     $\mathbf{70.0}$ &              $\mathbf{55.1 \pm 2.4}$ &     $\mathbf{32.7}$ &              $\mathbf{52.5 \pm 2.5}$ &  $\mathbf{46.5}$ &         $\mathbf{65.7 \pm 2.4}$ &       $\mathbf{34.4}$ &                $\mathbf{76.6 \pm 2.1}$ &  $\mathbf{54.4}$ \\
& STGCN$_{F}$          & $39.0 \pm 2.5$ &  $14.4$ &        $66.7 \pm 2.3$ &     $60.1$ &              $45.1 \pm 2.4$ &     $21.1$ &              $43.1 \pm 2.5$ &  $34.9$ &         $60.0 \pm 2.5$ &       $29.2$ &                $71.3 \pm 2.3$ &  $47.5$ \\
\hdashline
& 3DCNN         &  $46.0 \pm 2.5$ &  $12.8$ &        $64.9 \pm 2.4$ &     $52.0$ &              $10.8 \pm 1.5$ &     $13.6$ &              $32.0 \pm 2.3$ &  $19.3$ &         $45.9 \pm 2.5$ &       $14.7$ &                $71.6 \pm 2.3$ &  $46.3$  \\
\hline
\hline
     \end{tabular}
     }
    \caption{Accuracy ($A.$) and per-class averaged accuracy ($\overline{A}$) of various models on the test sets of the six tasks. For accuracy, we report the error margin as a confidence interval at $\alpha=0.05$ using asymptotic normal approximation. We omit error margins for balanced accuracy as the low number of classes results in a small sample size. Additional performance measures are reported in the appendix.}
    \label{tab:results}
\end{table*}

\subsection{Experimental Setup}
For each phonological property we generate dataset splits and train dedicated models separately. 
While a multi-class multi-label approach could achieve higher scores, by relying on potential interdependencies of different properties, we chose to model the properties in isolation, to disentangle the factors that affect the learnability of each property. From now on, when we mention the \textit{dataset}, we refer to an instance of the WLASL-Lex 2001 dataset, where labels are the values of a single phonological class.

We make this distinction because we produce six different train, validation and test splits (with a $70:15:15$ ratio) stratifying on the corresponding phonological property (\emph{Phoneme}). By doing so, we make sure that \emph{(a)} all splits contain all possible labels for a classification target (i.e. phonological property) and \emph{(b)} follow the same distribution. Since we source the videos from WLASL, we have multiple videos representing each gloss, therefore, randomly splitting our data will result in the fact that glosses in the test set might appear in the training set as well, signed by a different speaker. Thus, to investigate how well the models can predict properties on unseen glosses, we also produce label-stratified splits on gloss-level (\emph{Gloss}), such that videos of glosses in the validation and test set do not appear in training data and vice versa.
Thus, to summarise, experiments in the \emph{Phoneme} setting aim to evaluate the capability to recognise phonological properties of signs that were already encountered in the training data, but are performed by a different speaker in the test set. Conversely, experiments in the \emph{Gloss} setting aim to evaluate the capability to recognise phonological properties of signs completely \emph{unseen during training}.

We use an I3D model that has been pre-trained on Kinetics-400~\cite{i3d} and fine-tune it on raw videos from our datasets. 
The other models are trained from scratch using keypoints as input. We fix the length of all input to 150 frames, longer sequences are truncated while shorter sequences are looped to reach the fixed length.
We select the best performing model based on performance on the validation set and for the final test set performance we train the models on both train and validation sets. For more details on model selection, consult the appendix. We measure both accuracy, to investigate how well models perform in general, and class-balanced accuracy to take into account how well they are able to model different classes of the phonological properties.

\section{Results and discussion}






The upper half of Table~\ref{tab:results} presents the results for the six dataset splits for the \emph{Phoneme} setting, where glosses in test data could have appeared in training data as well. 
The poor performance of the simple MLP architecture suggests that the tasks are in fact challenging and do not exhibit easily exploitable regularities. Due to its simplicity, it is barely able to reach the baseline for some properties ($34\%$ vs. $35\%$ and $44\%$ vs. $50\%$ for \textit{movement} and \textit{flexion} respectively). In particular, MLP classifying based on FrankMocap  (MLP$_F$) output is often the worst performing combination. Conversely, STGCN using HRNet output (STGCN$_H$) outperforms other models on all six tasks. In some cases, for example when predicting \textit{movement} or \textit{flexion}, it is the only model which significantly surpasses the majority class baseline. This superior performance is expected, as this specific combination of the STGCN operating over HRNet-extracted keypoints has been shown to be the largest contributor to the SLR performance on the WLASL2000 dataset \cite{Jiang2021SignEnsemble}.

Models that operate over structured input often outperform the 3D CNN, demonstrating the utility of additional information provided by the skeleton features. The results also suggest that models using the HRNet skeleton output outperform those who use FrankMocap, possibly due to the confidence scores produced by HRNet and associated with the coordinates. This difference in performance suggests to conduct a more rigorous study to investigate the impact of different feature extraction methods as a possible future research direction.

The lower half of Table~\ref{tab:results} shows the performance of models to predict the phonological properties of unseen glosses (\emph{Gloss}). The performance of all tasks and all models deteriorates, suggesting that their success is partly derived from exploiting the similarities between glosses that appear in training and test data. However, the best model, STGCN$_H$, performs comparably to the \emph{Phoneme}-split, with a drop of less than $10$ accuracy points for five of the six tasks.

Often, crowd sourced \cite{ghioca} or automatically constructed datasets such as ours, have a performance ceiling, possibly due to incorrectly assigned ground truth labels or low quality of input data \cite{Chen2016,Schlegel2020}. To investigate the former, %
we measure the agreement on videos that all models misclassify using Fleiss' $\kappa$. Intuitively, if models consistently agree on a label different than the ground truth, the ground truth label might be wrong. We find that averaged across the six tasks, the agreement is negligible: $0.09 \pm 0.06$ and $0.11 \pm 0.09$ for \emph{Phoneme} and \emph{Gloss} split, respectively. 

Similarly, for the latter, if all models consistently fail to assign any correct label for a given video (e.g. all models err on a video appearing in the test sets of \emph{movement} and \emph{flexion}), this can hint at low quality of the input, making it impossible to predict anything correctly.
We find that this is not the case with WLASL-LEX2001, as videos appearing in test sets of different tasks tend to have a low mutual misclassification rate: $1\%$ and $0.7\%$ of videos appearing in test sets of two and three tasks were misclassified by all models for all associated tasks for the \emph{Phoneme} split. For the \emph{Gloss} split the numbers are $3$ and $0\%$ for two and three tasks, respectively. Together, these observations suggest that the models presented in this paper are unlikely to reach the performance ceiling on WLASL-Lex2001 and more advanced approaches could obtain even higher accuracy scores. 

\section{Conclusion}

In this paper, we discuss the task of Phonological Property Recognition (PPR). We automatically construct a dataset for the task featuring six phonological properties and analyse it extensively. We find that there is potential for improvement over our presented data-driven baseline approaches. Researchers pursuing this direction can focus on developing better-performing models, for example by relying on jointly learning all properties, as labels for different properties can be mutually dependent. 

Another possible avenue is to investigate the feasibility of using PRR to perform \emph{tokenisation} of continuous sign language speech, by decomposing it into multiple phonemes, which is identified as one of the big challenges of SLP \cite{Yin2021IncludingProcessing}.




\section*{Acknowledgements}
The authors would like to acknowledge the use of the Computational Shared Facility at The University of Manchester. The work was partially supported by the UKRI TAS Node on Trust, the US Air Force project THRIVE++ and the H2020 projects TRAINCREASE, eLADDA and PERSEO.

\bibliography{references, references_fede}
\bibliographystyle{acl_natbib}

\newpage
\appendix

\clearpage
\section{Hyperparameters optimization}
\label{sec:appendix}

Table~\ref{tab:hpopt} contains all the hyperparameters explored during our experiment over each different model. The best model is the one that maximises the Matthew's correlation coefficient
\newline \newline
\resizebox{\columnwidth}{!}{%
   $MCC = \frac{TP \cdot TN - FP \cdot FN}{\sqrt{(TP + FP)(TP + FN)(TN + FP)(TN + FN)}}$
}
\newline \newline
with $TP, TN, FP, FN$ being true/false positive/negative.
For the STGCN we use hyperparameters chosen by~\newcite{Jiang2021SignEnsemble}, because initial experiments on our data showed a difference of at most 2\% accuracy, which is within the uncertainty estimate. 
To find the optimal hyperparameters for the other models, we perform Bayesian optimisation over a pre-defined set. 
We maximise Matthews correlation coefficient (MCC)~\cite{MATTHEWS1975442} on the validation sets of all six tasks. We choose MCC as it provides a good trade-off between overall and class-level accuracy which is necessary due to the unbalance inherently present in our dataset.

\begin{table}[h]
\centering
\begin{tabular}{ll}
\textbf{Model} & \textbf{Parameters}                                                                                         \\ \hline \hline
MLP   & \begin{tabular}[c]{@{}l@{}}number of layers\\ hidden dimension\\ dropout\\ learning rate \\ scheduler step size\\ gamma\end{tabular}                  \\ \hline
RNN            & \begin{tabular}[c]{@{}l@{}}number of RNN layers\\ RNN hidden dimension\\ RNN dropout\end{tabular}           \\ \hline
STGCN & \begin{tabular}[c]{@{}l@{}}learning rate\\ number of groups\\ block size, \\ window size\\ scheduler step size\\ dropout\\ warmup epochs\end{tabular} \\ \hline
3D CNN         & \begin{tabular}[c]{@{}l@{}}dropout\\ learning rate\\ gamma\\ scheduler step size\\ window size\end{tabular} \\ \hline \hline
\end{tabular}%
\caption{Set of explored hyperparameters for each different model}
\label{tab:hpopt}
\end{table}

\section{Seed dependency}

Table~\ref{tab:seedep} illustrates the performance on the test set for each model with respect to chance as measured by training 5 models from different random seeds. The performance difference is negligible suggesting that model training is largely stable with regard to chance.

\begin{table}[h]
\centering
\begin{tabular}{ll}
\textbf{Model} & \textbf{Accuracy} \\ \hline \hline
MLP            & $74.39 \pm 0.35$    \\ \hline
RNN            & $79.12 \pm 0.46$    \\ \hline
STGCN          & $84.12 \pm 0.29$    \\ \hline
3D CNN         & $69.23 \pm 0.93$    \\ \hline \hline
\end{tabular}%
\caption{Mean and standard deviation of accuracy of all architectures trained with the HRNet output, measured on the \textsc{SignType} test set and averaged over 5 different random seeds. Results for the 3D CNN are obtained from the validation set.}
\label{tab:seedep}
\end{table}

\section{Phonological classes description}

\Crefrange{tab:selfing}{tab:movement} describe in detail the meaning of values for all the phonological classes according to ASL-Lex~\cite{Caselli2017}.

The cardinality is calculated on WLASL-Lex, which is why some classes that are in ASL-Lex are not represented (i.e., cardinality equal to 0).

\section{Additional results}

Table~\ref{tab:results-full} illustrates additional results for several different metrics. In particular, we report micro- and macro precision/recall and Matthews correlation coefficient. These metrics help to give a better understanding of the classification results, as they are affected more by data imbalance when compared to accuracy.

\begin{table*}[h]
\centering
\begin{tabular}{llc}
\textbf{Value} & \textbf{Definition}               & \textbf{Cardinality} \\ \hline
imrp           & index, middle, ring, pinky finger & 4824                 \\ \hline
imr            & index, middle, ring finger        & 95                   \\ \hline
mrp            & middle, ring, pinky finger        & 28                   \\ \hline
im             & index, middle finger              & 1296                 \\ \hline
ip             & index, pinky finger               & 51                   \\ \hline
mr             & middle, ring finger               & 0                    \\ \hline
mp             & middle, pinky finger              & 0                    \\ \hline
rp             & ring, pinky finger                & 0                    \\ \hline
i              & index finger                      & 2547                 \\ \hline
m              & middle finger                     & 259                  \\ \hline
r              & ring finger                       & 0                    \\ \hline
p              & pinky                             & 407                  \\ \hline
thumb          & thumb                             & 510                  \\ \hline
\end{tabular}%
\caption{Values and relative definitions for selected fingers}
\label{tab:selfing}
\end{table*}

\begin{table*}[h]
\centering
\begin{tabular}{llc}
\textbf{Value} & \textbf{Definition}                                          & \textbf{Cardinality} \\ \hline
Head           & Sign is produced on or near the head                         & 3137                 \\ \hline
Arm            & Sign is produced on or near the arm                          & 219                  \\ \hline
Body           & Sign is produced on or near the trunk                        & 1019                 \\ \hline
Hand           & Sign is produced on or near the non-dominant hand            & 2194                 \\ \hline
Neutral        & Sign is not produced in another location on the body         & 3448                 \\ \hline
Other          & Sign is produced in another unspecified location on the body & 0                    \\ \hline
\end{tabular}%
\caption{Values and relative definitions for major location}
\label{tab:majorloc}
\end{table*}

\begin{table*}[h]
\centering
\begin{tabular}{llc}
\textbf{Value} & \textbf{Definition}                                              & \textbf{Cardinality} \\ \hline
1              & Fully open: no joints of selected fingers are flexed             & 5037                 \\ \hline
2              & Bent (closed): non-base joints are flexed                        & 693                  \\ \hline
3              & Flat-open: base joints flexed less than 90 degrees               & 909                  \\ \hline
4              & Flat-closed: base joints flexed equal to or more that 90 degrees & 507                  \\ \hline
5              & Curved open: base and non-base joints flexed without contact     & 1130                 \\ \hline
6              & Curved closed: base and non-base joints flexed with contact      & 642                  \\ \hline
7              & Fully closed: base and non-base joints fully flexed              & 795                  \\ \hline
Stacked        & Stacked: Flexion of selected fingers differs                     & 123                  \\ \hline
Crossed        & Crossed                                                          & 181                  \\ \hline
\end{tabular}%
\caption{Values and relative definitions for flexion}
\label{tab:flexion}
\end{table*}

\begin{table*}[h]
\centering
\resizebox{\textwidth}{!}{%
\begin{tabular}{llc}
\textbf{Value} & \textbf{Definition}                                     & \textbf{Cardinality} \\ \hline
HeadTop        & Sign is produced on top of the head                     & 20                   \\ \hline
Forehead       & Sign is produced at the forehead                        & 246                  \\ \hline
Eye            & Sign is produced near the eye                           & 616                  \\ \hline
CheekNose      & Sign is produced on the cheek or nose                   & 511                  \\ \hline
UpperLip       & Sign is produced on the upper lip                       & 53                   \\ \hline
Mouth          & Sign is produced on the mouth                           & 431                  \\ \hline
Chin           & Sign is produced on the chin                            & 717                  \\ \hline
UnderChin      & Sign is produced under the chin                         & 74                   \\ \hline
UpperArm       & Sign is produced on the upper arm                       & 39                   \\ \hline
ElbowFront     & Sign is produced in the crook of the elbow              & 0                    \\ \hline
ElbowBack      & Sign is produced on the outside of the elbow            & 13                   \\ \hline
ForearmBack    & Sign is produced on the outside of the forearm          & 32                   \\ \hline
ForearmFront   & Sign is produced on the inside of the forearm           & 10                   \\ \hline
ForearmUlnar   & Sign is produced on the ulnar side of the forearm       & 56                   \\ \hline
WristBack      & Sign is produced on the back of the wriset              & 23                   \\ \hline
WristFront     & Sign is produced on the front of the wrist              & 0                    \\ \hline
Neck           & Sign is produced on the neck                            & 68                   \\ \hline
Shoulder       & Sign is produced on the shoulder                        & 101                  \\ \hline
Clavicle       & Sign is produced on the clavicle                        & 419                  \\ \hline
TorsoTop       & Sign is produced in the upper third of the torso        & 0                    \\ \hline
TorsoMid       & Sign is produced in the middle third of the torso       & 0                    \\ \hline
TorsoBottom    & Sign is produced in the bottom third of the torso       & 19                   \\ \hline
Waist          & Sign is produced at the waist                           & 34                   \\ \hline
Hips           & Sign is produced on the hips                            & 59                   \\ \hline
Palm           & Sign is produced on the plam of the non-dominant hand   & 925                  \\ \hline
FingerFront  & Sign is produced on the front of the fingers of the non-dominant hand & 99  \\ \hline
PalmBack     & Sign is produced on the back of the palm of the non-dominant hand     & 218 \\ \hline
FingerBack   & Sign is produced on the back of the fingers of the non-dominant hand  & 186 \\ \hline
FingerRadial & Sign is produced on the radial side of the non-dominant hand          & 410 \\ \hline
FingerUlnar  & Sign is produced on the ulnar side of the non-dominant hand           & 40  \\ \hline
FingerTip    & Sign is produced on the tip of the fingers of the non-dominant hand   & 158 \\ \hline
Heel           & Sign is produced on the heel of the non-dominant hand   & 88                   \\ \hline
Other          & Sign is produced in an unspecified location on the body & 707                  \\ \hline
Neutral        & Sign is not produced on or near the body                & 3390                 \\ \hline
\end{tabular}%
}
\caption{Values and relative definitions for minor location}
\label{tab:minorloc}
\end{table*}

\begin{table*}[h]
\centering
\resizebox{\textwidth}{!}{%
\begin{tabular}{llc}
\textbf{Value} & \textbf{Definition}                                        & \textbf{Cardinality} \\ \hline
One Handed     & Sign only recruits one hand                                & 3939                 \\ \hline
\begin{tabular}[c]{@{}l@{}}Symmetrical\\ Or Alternating\end{tabular} &
  \begin{tabular}[c]{@{}l@{}}Sign recruits both hands\\ Phonological specifications for both hands are identical\\ Movement of both hands is either symmetrical or alternating\end{tabular} &
  3358 \\ \hline
\begin{tabular}[c]{@{}l@{}}Asymmetrical\\ Same Handshape\end{tabular} &
  \begin{tabular}[c]{@{}l@{}}Sign recruits both hands\\ Only the dominant hand moves\\ The location and orientation of the hands may differ,\\ but the other specifications of handshape are the same\\ Non-Dominant hand must be an unmarked handshape (B A S 1 C O 5)\end{tabular} &
  938 \\ \hline
\begin{tabular}[c]{@{}l@{}}Asymmetrical\\ Different Handshape\end{tabular} &
  \begin{tabular}[c]{@{}l@{}}Sign recruits both hands\\ Only the dominant hand moves\\ The location and orientation of the hands may differ,\\ and the other specifications of handshape are not the same\\ Non-Dominant hand must be an unmarked handshape (B A S 1 C O 5)\end{tabular} &
  1639 \\ \hline
Other          & Sign violates Battison's Symmetry and Dominance Conditions & 143                  \\ \hline
\end{tabular}%
}
\caption{Values and relative definitions for sign type}
\label{tab:signtype}
\end{table*}

\begin{table*}[h]
\centering
\begin{tabular}{llc}
\textbf{Value} & \textbf{Definition}                                                & \textbf{Cardinality} \\ \hline
Straight       & Straight movement of the dominant hand through xyz space           & 1938                 \\ \hline
Curved &
  \begin{tabular}[c]{@{}l@{}}Single arc movement of the dominant hand through xyz space\\ Hands may or may not make contact with multiple locations\end{tabular} &
  1255 \\ \hline
BackAndForth   & Sequence of more than one straight or curved movements             & 3549                 \\ \hline
Circular &
  \begin{tabular}[c]{@{}l@{}}Circular movement of the dominant hand through space\\ Rotation alone does not constitute a circular movement\end{tabular} &
  1129 \\ \hline
None           & Entire sign (or first free morpheme) does not have a path movement & 1748                 \\ \hline
Other          & Sign has another unspecified path movement                         & 398                  \\ \hline
\end{tabular}%
\caption{Values and relative definitions for movement}
\label{tab:movement}
\end{table*}

\begin{sidewaystable}[!t]
    \centering
    \resizebox{.99\textwidth}{!}{%
    \begin{tabular}{p{0.05cm} l | c G c G c G c G c G c G c G c G c G c G c G c G c G c G c G}
    &  & \multicolumn{5}{c}{\textbf{\textsc{Flexion}}} &  \multicolumn{5}{c}{\textbf{\textsc{MajLocation}}} &  \multicolumn{5}{c}{\textbf{\textsc{MinLocation}}}  &  \multicolumn{5}{c}{\textbf{\textsc{Movement}}}  &  \multicolumn{5}{c}{\textbf{\textsc{Fingers}}}  &  \multicolumn{5}{c}{\textbf{\textsc{Signtype}}}  \\
    &  & $P_\mu$ & $P_{M}$ & $R_\mu$ & $R_M$ & $MCC$ & 
    $P_\mu$ & $P_{M}$ & $R_\mu$ & $R_M$ & $MCC$ &
    $P_\mu$ & $P_{M}$ & $R_\mu$ & $R_M$ & $MCC$ & 
    $P_\mu$ & $P_{M}$ & $R_\mu$ & $R_M$ & $MCC$ & 
    $P_\mu$ & $P_{M}$ & $R_\mu$ & $R_M$ & $MCC$ & 
    $P_\mu$ & $P_{M}$ & $R_\mu$ & $R_M$ & $MCC$ \\
        \hline
        \hline
        \multirow{8}{*}{\rotatebox[origin=c]{90}{\emph{Phoneme}}} &
Baseline & $50.3$ & $5.59$ & $50.3$ & $11.11$ & $0.0$ & $34.4$ & $6.88$ & $34.4$ & $20.0$ & $0.0$ & $33.87$ & $1.06$ & $33.87$ & $3.12$ & $0.0$ & $35.46$ & $5.91$ & $35.46$ & $16.67$ & $0.0$ & $48.17$ & $5.35$ & $48.17$ & $11.11$ & $0.0$ & $39.32$ & $7.86$ & $39.32$ & $20.0$ & $0.0$  \\
& MLP$_{H}$ &   $44.1$ &  $24.5$ &  $44.1$ &  $20.7$ &  $14.6$ &  $70.3$ &  $65.8$ &  $70.3$ &  $64.0$ &  $58.9$ &  $51.6$ &  $37.3$ &  $51.6$ &  $28.2$ &  $41.6$ &  $34.5$ &  $28.0$ &  $34.5$ &  $26.9$ &  $13.9$ &  $59.5$ &  $29.6$ &  $59.5$ &  $25.0$ &  $37.7$ &  $73.9$ &  $54.1$ &  $73.9$ &  $52.6$ &  $62.5$ \\
& MLP$_{F}$ &     $50.3$ &  $5.6$ &   $50.3$ &  $11.1$ &  $0.9$ &   $57.8$ &  $52.3$ &  $57.8$ &  $46.8$ &  $41.2$ &  $34.3$ &  $13.9$ &  $34.3$ &  $9.1$ &   $17.9$ &  $34.3$ &  $13.1$ &  $34.3$ &  $18.7$ &  $5.7$ &   $43.4$ &  $17.5$ &  $43.4$ &  $12.9$ &  $4.6$ &   $67.1$ &  $38.1$ &  $67.1$ &  $42.8$ &  $52.7$ \\
& RNN$_{H}$ &   $49.0$ &  $32.1$ &  $49.0$ &  $30.0$ &  $25.4$ &  $75.8$ &  $75.2$ &  $75.8$ &  $72.4$ &  $66.4$ &  $64.3$ &  $54.3$ &  $64.3$ &  $46.0$ &  $57.4$ &  $35.1$ &  $30.1$ &  $35.1$ &  $29.5$ &  $15.9$ &  $71.0$ &  $53.3$ &  $71.0$ &  $46.5$ &  $56.6$ &  $78.7$ &  $61.2$ &  $78.7$ &  $58.8$ &  $69.4$ \\
& RNN$_{F}$ &     $50.3$ &  $5.6$ &   $50.3$ &  $11.1$ &  $0.0$ &   $63.9$ &  $56.7$ &  $63.9$ &  $52.2$ &  $50.1$ &  $30.3$ &  $4.7$ &   $30.3$ &  $4.0$ &   $4.9$ &   $35.4$ &  $21.4$ &  $35.4$ &  $18.1$ &  $5.2$ &   $46.5$ &  $9.2$ &   $46.5$ &  $12.4$ &  $8.5$ &   $70.9$ &  $60.6$ &  $70.9$ &  $46.8$ &  $58.3$ \\
& GTN${_H}$ &   $62.4$ &  $55.4$ &  $62.4$ &  $45.0$ &  $43.9$ &  $83.2$ &  $80.6$ &  $83.2$ &  $78.6$ &  $76.8$ &  $74.5$ &  $66.7$ &  $74.5$ &  $63.5$ &  $69.8$ &  $63.6$ &  $62.1$ &  $63.6$ &  $58.2$ &  $52.7$ &  $73.8$ &  $71.7$ &  $73.8$ &  $56.0$ &  $61.1$ &  $84.5$ &  $74.9$ &  $84.5$ &  $69.6$ &  $77.7$ \\
& GTN$_{F}$ &     $43.4$ &  $23.6$ &  $43.4$ &  $20.8$ &  $15.3$ &  $70.5$ &  $66.4$ &  $70.5$ &  $62.1$ &  $58.9$ &  $53.0$ &  $43.9$ &  $53.0$ &  $40.0$ &  $43.8$ &  $45.7$ &  $40.8$ &  $45.7$ &  $37.8$ &  $28.6$ &  $63.1$ &  $39.0$ &  $63.1$ &  $32.8$ &  $44.3$ &  $73.0$ &  $56.8$ &  $73.0$ &  $53.1$ &  $61.1$ \\
\hdashline
& 3DCNN &  $46.5$ &  $17.8$ &  $46.5$ &  $13.2$ &  $5.4$ &   $64.3$ &  $57.2$ &  $64.3$ &  $55.2$ &  $50.3$ &  $42.3$ &  $22.8$ &  $42.3$ &  $18.6$ &  $29.1$ &  $32.9$ &  $23.4$ &  $32.9$ &  $20.8$ &  $7.5$ &   $47.5$ &  $17.8$ &  $47.5$ &  $14.5$ &  $14.6$ &  $69.5$ &  $44.9$ &  $69.5$ &  $44.8$ &  $55.6$ \\
        \hline
        \hline
        \multirow{6}{*}{\rotatebox[origin=c]{90}{\emph{Gloss}}} &
Baseline & $53.03$ & $5.89$ & $53.03$ & $11.11$ & $0.0$ & $35.69$ & $7.14$ & $35.69$ & $20.0$ & $0.0$ & $42.03$ & $2.1$ & $42.03$ & $5.0$ & $0.0$ & $35.21$ & $5.87$ & $35.21$ & $16.67$ & $0.0$ & $47.38$ & $5.92$ & $47.38$ & $12.5$ & $0.0$ & $38.28$ & $7.66$ & $38.28$ & $20.0$ & $0.0$  \\
& MLP$_{H}$ &   $44.6$ &  $18.6$ &  $44.6$ &  $15.5$ &  $8.3$ &   $68.1$ &  $62.0$ &  $68.1$ &  $56.6$ &  $55.5$ &  $47.3$ &  $16.8$ &  $47.3$ &  $13.1$ &  $32.5$ &  $28.4$ &  $20.4$ &  $28.4$ &  $19.8$ &  $4.9$ &   $56.2$ &  $21.4$ &  $56.2$ &  $20.3$ &  $32.0$ &  $75.3$ &  $50.6$ &  $75.3$ &  $50.7$ &  $64.3$ \\
& MLP$_{F}$ &     $52.8$ &  $5.9$ &   $52.8$ &  $11.1$ &  $-2.1$ &  $56.7$ &  $46.2$ &  $56.7$ &  $42.9$ &  $39.5$ &  $38.3$ &  $11.7$ &  $38.3$ &  $7.9$ &   $18.1$ &  $37.1$ &  $15.9$ &  $37.1$ &  $21.7$ &  $12.5$ &  $39.3$ &  $10.4$ &  $39.3$ &  $11.1$ &  $0.4$ &   $68.4$ &  $37.7$ &  $68.4$ &  $41.2$ &  $54.3$ \\
& RNN$_{H}$ &   $39.6$ &  $19.8$ &  $39.6$ &  $18.0$ &  $10.9$ &  $72.8$ &  $68.0$ &  $72.8$ &  $67.3$ &  $62.4$ &  $49.3$ &  $19.6$ &  $49.3$ &  $17.5$ &  $36.7$ &  $32.2$ &  $25.7$ &  $32.2$ &  $24.9$ &  $11.3$ &  $60.7$ &  $36.9$ &  $60.7$ &  $32.5$ &  $40.3$ &  $75.4$ &  $55.0$ &  $75.4$ &  $53.5$ &  $64.6$ \\
& RNN$_{F}$ &     $53.0$ &  $5.9$ &   $53.0$ &  $11.1$ &  $0.0$ &   $64.1$ &  $57.3$ &  $64.1$ &  $52.6$ &  $50.5$ &  $44.4$ &  $15.1$ &  $44.4$ &  $12.3$ &  $27.9$ &  $36.7$ &  $11.2$ &  $36.7$ &  $20.1$ &  $10.0$ &  $27.3$ &  $10.6$ &  $27.3$ &  $12.7$ &  $3.0$ &   $72.0$ &  $41.3$ &  $72.0$ &  $46.9$ &  $60.4$ \\
& GTN$_{H}$ &   $49.1$ &  $25.6$ &  $49.1$ &  $21.6$ &  $18.9$ &  $77.3$ &  $72.1$ &  $77.3$ &  $70.0$ &  $68.6$ &  $55.1$ &  $25.1$ &  $55.1$ &  $23.3$ &  $43.4$ &  $52.5$ &  $49.4$ &  $52.5$ &  $46.5$ &  $38.0$ &  $65.7$ &  $37.2$ &  $65.7$ &  $30.6$ &  $47.8$ &  $76.6$ &  $54.9$ &  $76.6$ &  $54.4$ &  $66.2$ \\
& GTN$_{F}$ &     $39.0$ &  $15.1$ &  $39.0$ &  $14.4$ &  $4.7$ &   $66.7$ &  $63.2$ &  $66.7$ &  $60.1$ &  $53.9$ &  $45.1$ &  $15.7$ &  $45.1$ &  $13.2$ &  $31.1$ &  $43.1$ &  $36.0$ &  $43.1$ &  $34.9$ &  $25.8$ &  $60.0$ &  $32.5$ &  $60.0$ &  $29.2$ &  $39.4$ &  $71.3$ &  $47.6$ &  $71.3$ &  $47.5$ &  $58.5$ \\
\hdashline
& 3DCNN &  $46.0$ &  $12.0$ &  $46.0$ &  $12.8$ &  $4.5$ &   $65.0$ &  $57.5$ &  $65.0$ &  $52.0$ &  $51.8$ &  $10.8$ &  $12.0$ &  $10.8$ &  $9.7$ &   $9.5$ &   $32.0$ &  $18.7$ &  $32.0$ &  $19.3$ &  $6.0$ &   $45.9$ &  $15.1$ &  $45.9$ &  $14.7$ &  $10.7$ &  $71.6$ &  $46.3$ &  $71.6$ &  $46.3$ &  $58.7$ \\
\hline  
\hline
     \end{tabular}
     }
    \caption{Micro-averaged ($\mu$), macro-averaged ($M$) precision ($P$) and recall ($R$) and Matthews correlation coefficient ($MCC$) of various models on the test sets of the six tasks. We omit error margins as the low number of classes results in a small sample size.}
    \label{tab:results-full}
\end{sidewaystable}

\end{document}